# Representing Bayesian Networks within Probabilistic Horn Abduction

David Poole
Department of Computer Science,
University of British Columbia,
Vancouver, B.C., Canada V6T 1W5
poole@cs.ubc.ca

## Abstract

This paper presents a simple framework for Horn-clause abduction, with probabilities associated with hypotheses. It is shown how this representation can represent any probabilistic knowledge representable in a Bayesian belief network. The main contributions are in finding a relationship between logical and probabilistic notions of evidential reasoning. This can be used as a basis for a new way to implement Bayesian Networks that allows for approximations to the value of the posterior probabilities, and also points to a way that Bayesian networks can be extended beyond a propositional language.

## 1 Introduction

In this paper we pursue the idea of having a logical language that allows for a pool of possible hypotheses [Poole et al., 1987; Poole, 1988], with probabilities associated with the hypotheses [Neufeld and Poole, 1987]. We choose a very simple logical language with a number of assumptions on the structure of the knowledge base and independence amongst hypotheses. This is intended to reflect a compromise between simplicity to implement and representational adequacy. To show that these assumptions are not unreasonable, and to demonstrate representational adequacy, we show how arbitrary Bayesian networks can be represented by the formalism. The main contributions of this embedding are:

- It shows a relationship between logical and probabilistic notions of evidential reasoning. In particular it provides some evidence for the use of abduction and assumption-based reasoning as the logical analogue of the independence of Bayesian Networks. In earlier work [Poole, 1989; Poole, 1990] a form of assumption-based reasoning where we abduce to causes and then makes assumptions in order to predict what should follow is developed; it is a similar mechanism that is used here to characterise Bayesian networks.

- It gives a different way to implement Bayesian networks[1]. The main advantage of this implementation is that it gives a way to approximate the probability with a known error bound.

- Because the underlying language is not propositional, it gives us a way to extend Bayesian networks to a richer language. This corresponds to a form of dynamic construction of Bayesian networks [Horsch and Poole, 1990].

In [Poole, 1991], it is argued that the probabilistic Horn abduction framework provides a good compromise between representational and heuristic adequacy for diagnostic tasks. It showed how the use of variables can be used to extend the purely propositional diagnostic frameworks, and how the use of probabilities can be naturally used to advantage in logical approaches to diagnosis.

## 2 Probabilistic Horn Abduction

### 2.1 Horn clause abduction

The formulation of abduction used is a simplified form of Theorist [Poole et al., 1987; Poole, 1988]. It is simplified in being restricted to Horn clauses.

Although the idea [Neufeld and Poole, 1987] is not restricted to Horn clauses (we could extend it to disjunction and classical negation [Poole et al., 1987] or to negation as failure [Eshghi and Kowalski, 1989]), in order to empirically test the framework, it is important to find where the simplest representations fail. It may be the case that we need the extra representational power of more general logics; we can only demonstrate this by doing without the extra representational power.

We use the normal Prolog definition of an atomic sym-

---

[1] For composite beliefs this is closely related to the algorithm of Shimony and Charniak [1990], but also is developed for finding posterior probabilities of hypotheses (section 3.2)

bol [Lloyd, 1987]. A Horn clause is of the form:

$$a.$$
$$a \leftarrow a_1 \wedge ... \wedge a_n.$$
$$false \leftarrow a_1 \wedge ... \wedge a_n.$$

where $a$ and each $a_i$ are atomic symbols. *false* is a special atomic symbol that is not true in any interpretation[2].

An abductive scheme is a pair $\langle F, H \rangle$ where

- $F$ is a set of Horn clauses. Variables in $F$ are implicitly universally quantified.

- $H$ is a set of atoms, called the "assumables" or the "possible hypotheses". Associated with each assumable is a prior probability.

Here (and in our implementation) we write

$$assumable(h, p).$$

where $h$ is a (possibly open) atom, and $p$ is a number $0 \leq p \leq 1$ to mean that for every ground instance $h\theta$ of $h$, $h\theta \in H$ and $P(h\theta) = p$.

**Definition 2.1** [Poole *et al.*, 1987; Poole, 1988] If $g$ is a ground formula, an **explanation** of $g$ from $\langle F, H \rangle$ is a subset $D$ of $H$ such that

- $F \cup D \models g$ and

- $F \cup D \not\models false$.

The first condition says that, $D$ is a sufficient cause for *obs*, and the second says that $D$ is possible (i.e., $F \cup D$ is consistent).

A **minimal explanation** of $g$ is an explanation of $g$ such that no strict subset is an explanation of $g$.

## 2.2 Probabilities

Associated with each possible hypothesis is a prior probability. The aim is to compute the posterior probability of the minimal explanations given the observations. Abduction gives us what we want to compute the probability of and probability theory gives a measure over the explanations [Neufeld and Poole, 1987].

To compute the posterior probability of an explanation $D = \{h_1, ..., h_n\}$ of observation *obs* given observation *obs*, we use Bayes rule and the fact that $P(obs|D) = 1$

---

[2]Notice that we are using Horn clauses differently from how Prolog uses Horn clauses. In Prolog, the database consists of definite clauses, and the queries provide the negative clauses [Lloyd, 1987]. Here the database consists of definite and negative clauses, and we build a constructive proof of an observation.

as the explanation logically implies the observation:

$$P(D|obs) = \frac{P(obs|D) \times P(D)}{P(obs)}$$
$$= \frac{P(D)}{P(obs)}$$

The value, $P(obs)$ is the prior probability of the observation, and is a constant factor for all explanations. We compute the prior probability of the conjunction of the hypotheses using:

$$P(h_1 \wedge ... \wedge h_{n-1} \wedge h_n) = P(h_n|h_1 \wedge ... \wedge h_{n-1})$$
$$\times P(h_1 \wedge ... \wedge h_{n-1})$$

The value of $P(h_1 \wedge ... \wedge h_{n-1})$ forms a recursive call, with $P(true) = 1$. The only other thing that we need to compute is

$$P(h_n|h_1 \wedge ... \wedge h_{n-1})$$

If $h_n$ is inconsistent with the other hypotheses, then the above conditional probability is zero. These are the cases that are removed by the inconsistency requirement. If $h_n$ is implied by the other hypotheses, the probability should be one. This case never arises for minimal explanations.

While any method can be used to compute this conditional probability, we assume that the logical dependencies impose the only statistical dependencies on the hypotheses.

**Assumption 2.2** *Logically independent instances of hypotheses are probabilistically independent.*

**Definition 2.3** A set $D$ of hypotheses are **logically independent** (given $F$) if there is no $S \subset D$ and $h \in D \setminus S$ such that

$$F \cup S \models h \quad \text{or} \quad F \cup S \models \neg h$$

The assumptions in a minimal explanation are always logically independent.

Under assumption 2.2, if $\{h_1, ..., h_n\}$ are part of a minimal explanation, then

$$P(h_n|h_1 \wedge ... \wedge h_{n-1}) = P(h_n)$$

thus

$$P(h_1 \wedge ... \wedge h_n) = \prod_{i=1}^{n} P(h_i)$$

To compute the prior of the minimal explanation we multiply the priors of the hypotheses. The posterior probability of the explanation is proportional to this.

The justification for the reasonableness (and universality) of this assumption is based on Reichenbach's *principle of the common cause*:

"If coincidences of two events $A$ and $B$ occur more frequently than their independent occurrence, ... then there exists a common cause for these events ..." [Reichenbach, 1956, p. 163].

When there is a dependency amongst hypotheses, we invent a new hypothesis to explain that dependence. Thus the assumption of independence, while it gives a restriction on the knowledge bases that are legal, really gives no restriction on the domains that can be represented.

### 2.3 Relations between explanations

The remaining problem in the probabilistic analysis is in determining the value of $P(obs)$.

When using abduction we often assume that the diagnoses are covering. This can be a valid assumption if we have anticipated all eventualities, and the observations are within the domain of the expected observations (usually if this assumption is violated there are no explanations). This is also supported by recent attempts at a completion semantics for abduction [Poole, 1988; Console et al., 1989; Konolige, 1991]. The results show how abduction can be considered as deduction on the "closure" of the knowledge base that includes statements that the given causes are the only causes. The closure implies the observation are logically equivalent to the disjunct of its explanations. We make this assumption explicit here:

**Assumption 2.4** *The diagnoses are covering.*

For the probabilistic calculation we make an additional assumption:

**Assumption 2.5** *The diagnoses are disjoint (mutually exclusive).*

It turns out to be straight forward to ensure that these properties hold, for observations that we can anticipate[3]. We make sure that the rules for each possible subgoal are disjoint and covering. This can be done locally for each atom that may be part of an observation or used to explain an observation.

When building the knowledge base, we use the local property that the rules for a subgoal are exclusive and covering to ensure that the explanations generated are exclusive and covering.

Under these assumptions, if $\{e_1, ..., e_n\}$ is the set of all explanations of $g$:

$$\begin{aligned} P(g) &= P(e_1 \vee e_2 \vee ... \vee e_n) \\ &= P(e_1) + P(e_2) + ... + P(e_n) \end{aligned}$$

---

[3]Like other systems (e.g., [Pearl, 1988b]), we have to assume that unanticipated observations are irrelevant.

## 3 Representing Bayesian networks

In this section we give the relationship between Bayesian networks and our probabilistic Horn abduction. We show how any probabilistic knowledge that can be represented in a Bayesian network, can be represented in our formalism.

Suppose we have a Bayesian network with random variables $a_1, ..., a_n$, such that random variable $a_i$ can have values $v_{i,1}, ..., v_{i,n_i}$. We represent random variable $a_i$ having value $v_{i,j}$ as the proposition $a_i(v_{i,j})$.

The first thing we need to do is to state that the values of variables are mutually exclusive. For each $i$ and for each $j, k$ such that $j \neq k$, we have the rule

$$false \leftarrow a_i(v_{i,j}) \wedge a_i(v_{i,k})$$

A *Bayesian network* [Pearl, 1988b] is a directed acyclic network where the nodes represent random variables, and the arcs represent a directly influencing relation. An arc from variable $b$ to variable $a$ represents the fact that variable $b$ *directly influences* variable $a$; the relation *influences* is the transitive closure of the directly influences relation. *Terminal nodes* of a Bayesian network are those variables that do not influence any other variables. The *depth* of a node is the length of the longest (directed) path leading into the node. A *composite belief* [Pearl, 1987] is an assignment of a value to every random variable.

Suppose variable $a$ is directly influenced by variables $\Pi_a = b_1, ..., b_m$ (the "parents" of $a$) in a Bayesian network. The independence assumption embedded in a Bayesian Network [Pearl, 1988b] is given by

$$P(a|\Pi_a \wedge v) = P(a|\Pi_a)$$

where $v$ is a variable (or conjunction of variables) such that $a$ does not influence $v$ (or any conjunct in $v$).

The network is represented by a rule that relates a variable with its parents:

$$a(V) \leftarrow b_1(V_1) \wedge ... \wedge b_m(V_m) \wedge c\_a(V, V_1, ..., V_m)$$

The intended interpretation of $c\_a(V, V_1, ..., V_m)$ is that $a$ has value $V$ because $b_1$ has value $V_1$,..., and $b_m$ has value $V_m$.

Associated with the Bayesian network is a contingency table which gives the marginal probabilities of the values of $a$ depending on the values of $b_1, ..., b_m$. This will consist of probabilities of the form

$$P(a = v|b_1 = v_1, ..., b_m = v_m) = p$$

This is translated into the assertion

$$assumable(c\_a(v, v_1, v_2, ..., v_m), p).$$

Nodes with no parents can be just made assumable, with the appropriate probabilities (rather than inventing a new hypothesis and the above procedure would prescribe).

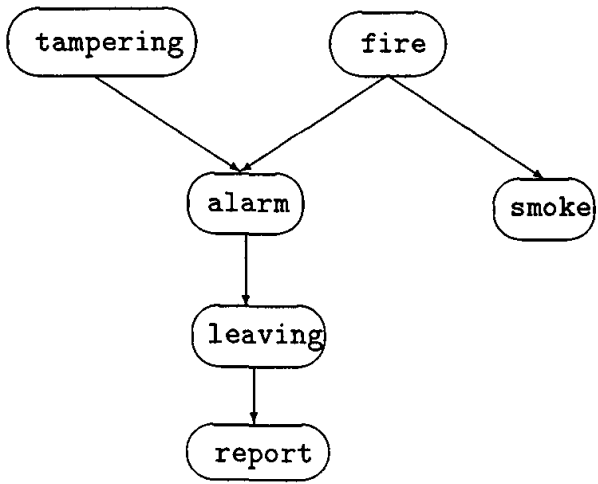

Figure 1: An influence diagram for a smoking alarm.

**Example 3.1** Consider a representation of the influence diagram of figure 3.1, with the following conditional probability distributions:

$$\begin{aligned}
p(fire) &= 0.01 \\
p(smoke|fire) &= 0.9 \\
p(smoke|\neg fire) &= 0.01 \\
p(tampering) &= 0.02 \\
p(alarm|fire \wedge tampering) &= 0.5 \\
p(alarm|fire \wedge \neg tampering) &= 0.99 \\
p(alarm|\neg fire \wedge tampering) &= 0.85 \\
p(alarm|\neg fire \wedge \neg tampering) &= 0.0001 \\
p(leaving|alarm) &= 0.88 \\
p(leaving|\neg alarm) &= 0.001 \\
p(report|leaving) &= 0.75 \\
p(report|\neg leaving) &= 0.01
\end{aligned}$$

The following is a representation of this Bayesian network in out formalism.

```
assumable( fire(yes), 0.01 ).
assumable( fire(no), 0.99 ).

false <- fire(yes), fire(no).

smoke(Sm) <- fire(Fi),
             c_smoke(Sm,Fi).
false <- smoke(yes),
         smoke(no).

assumable( c_smoke(yes,yes), 0.9 ).
assumable( c_smoke(no,yes), 0.1 ).
assumable( c_smoke(yes,no), 0.01 ).
assumable( c_smoke(no,no), 0.99 ).

assumable( tampering(yes), 0.02 ).
assumable( tampering(no), 0.98 ).

alarm(Al) <- fire(Fi), tampering(Ta),
             c_alarm(Al,Fi,Ta).
false <- alarm(yes),
         alarm(no).

assumable( c_alarm(yes,yes,yes), 0.50 ).
assumable( c_alarm(no,yes,yes),  0.50 ).
assumable( c_alarm(yes,yes,no),  0.99 ).
assumable( c_alarm(no,yes,no),   0.01 ).
assumable( c_alarm(yes,no,yes),  0.85 ).
assumable( c_alarm(no,no,yes),   0.15 ).
assumable( c_alarm(yes,no,no),   0.0001 ).
assumable( c_alarm(no,no,no),    0.9999 ).

leaving(Le) <- alarm(Al),
               c_leaving(Le,Al).
false <- leaving(yes),
         leaving(no).

assumable( c_leaving(yes,yes), 0.88 ).
assumable( c_leaving(no,yes), 0.12 ).
assumable( c_leaving(yes,no), 0.001 ).
assumable( c_leaving(no,no), 0.999 ).

report(Le) <- leaving(Al),
              c_report(Le,Al).
false <- report(yes),
         report(no).

assumable( c_report(yes,yes), 0.75 ).
assumable( c_report(no,yes), 0.25 ).
assumable( c_report(yes,no), 0.01 ).
assumable( c_report(no,no), 0.99 ).
```

### 3.1 Composite Beliefs

A composite belief [Pearl, 1987] is an assignment of a value to every random variable. The composite belief with the highest probability has also been called a MAP assignment [Charniak and Shimony, 1990]. These composite beliefs have been most used for diagnosis [de Kleer and Williams, 1987; de Kleer and Williams, 1989; Peng and Reggia, 1990] (see [Poole and Provan, 1990] for a discussion on the appropriateness of this).

**Lemma 3.2** *The minimal explanations of the terminal variables having particular values correspond to the composite beliefs in the Bayesian network with the terminals having those values. The priors for the explanations and the composite beliefs are identical.*

The proof of this lemma and for lemma 3.4 appears in appendix A.

As the same procedure can be used to get from the priors of composite hypotheses and explanations to the posteriors given some observations, the following theorem is a direct corollary of lemma 3.2.

**Theorem 3.3** *If the observed variables include all terminal variables, the composite beliefs with the observed variables having particular values correspond exactly to the explanations of the observations, and with the same posterior probability.*

If the observed variables do not include all terminal values, we need to decide what it is that we want the probability of [Poole and Provan, 1990]. If we want to commit to the value of all variables, we consider the set of possible observations that include assigning values to terminal nodes. That is, if $o$ was our observation that did not not include observing a value for variables $\bar{a}$, then we need to consider the observations $o \wedge \bar{a}(\bar{v})$, for each tuple $\bar{v}$ of values of variables $\bar{a}$. To find the accurate probabilities we need to normalise over the sum of all of the explanations. Whether or not we want to do this is debatable.

### 3.2 Posterior Probabilities of Propositions

In the previous section, the observations to be explained corresponded exactly to the conditioning variables. This corresponds to the use of "abduction" in [Poole, 1989]. In this section we show a relationship to the combination of abducing to causes and default reasoning to predictions from these causes [Poole, 1989; Poole, 1990].

Let $expl(\alpha)$ be the set of minimal explanations of proposition (or conjunction) $\alpha$. Define

$$\mathcal{M}(\alpha) = \sum_{E \in expl(\alpha)} P(E)$$

**Lemma 3.4** *If $H$ is a set of assignments to variables in a Bayesian Network, and $H'$ is the analogous propositions to $H$ in the corresponding probabilistic Horn abduction system, then*

$$P(H) = \mathcal{M}(H')$$

A simple corollary of the above lemma can be used to determine the posterior probability of a hypothesis based on some observations:

**Theorem 3.5**

$$P(x_i = v_i | obs) = \frac{\mathcal{M}(obs' \wedge x_i(v_i))}{\mathcal{M}(obs')}$$

The denominator can be obtained by finding the explanations of the observations. The numerators can be obtained by explaining $x_i(v_i)$ from these explanations.

## 4 Best-first abduction

We are currently experimenting with a number of implementations based on Logic programming technology or on ATMS technology. These are implemented by a branch and bound search where we consider the partial explanation with the least cost (highest probability) at any time.

The implementations keep a priority queue of sets of hypotheses that could be extended into explanations ("partial explanations"). At any time the set of all the explanations is the set of already generated explanations, plus those explanations that can be generated from the partial explanations in the priority queue.

Formally[4], a partial explanation is a pair

$$\langle g \leftarrow C, D \rangle$$

where $g$ is an atom, $C$ is a conjunction of atoms and $D$ is a set of hypotheses.

Initially the priority queue to explain $a$ contains

$$\{\langle a \leftarrow a, \{\} \rangle, \langle false \leftarrow false, \{\} \rangle\}$$

We thus try simultaneously try to find explanations of $a$ and "explanations" of $false$ (forming *nogoods* in ATMS terminology) that can be used to prune other partial explanations.

At each step we choose the element

$$\langle g \leftarrow C, D \rangle$$

of the priority queue with maximum prior probability of $D$, but when partial explanations are equal we have a preference for explanations of $false$.

We have an explanation when $C$ is the empty conjunction. Otherwise, suppose $C$ is conjunction $a \wedge R$.

There are two operations that can be carried out. The first is a form of SLD resolution [Lloyd, 1987], where for each rule

$$a \leftarrow b_1 \wedge ... \wedge b_n$$

in $F$, we generate the partial explanation

$$\langle g \leftarrow b_1 \wedge ... \wedge b_n \wedge R, D \rangle .$$

The second operation is used when $a \in H$. In this case we produce the partial explanation

$$\langle g \leftarrow R, \{a\} \cup D \rangle$$

This procedure, under reasonable assumptions, will find the explanations in order of liklihood.

---

[4]Here we give only the bare-bones of the goal-directed procedure; there is an analogous bottom-up procedure that we are also experimenting with. The analysis is similar for that procedure. We also only give the propositional version here. The lifting to the general case by the use of substitutions is straightforward [Lloyd, 1987].

It turns out to be straight forward to give an upper bound on the probability mass in the priority queue.

If $\langle g \leftarrow C, D \rangle$ is in the priority queue, then it can possibly be used to generate explanations $D_1, ..., D_n$. Each $D_i$ will be of the form $D \cup D_i'$. We can place a bound on the probability mass of all of the $D_i$, by

$$\begin{aligned} p(D_1 \vee ... \vee D_n) &= p(D \wedge (D_1' \vee ... \vee D_n')) \\ &\leq p(D) \end{aligned}$$

This means that we can put an bound on the range of probabilities of an goal based on finding just some of the explanations of the goal. Suppose we have goal $g$, and we have generated explanations $\mathcal{D}$. Let

$$P_\mathcal{D} = \sum_{D \in \mathcal{D}} P(D)$$

$$P_Q = \sum_{D : \langle g \leftarrow C, D \rangle \in Q} P(D)$$

where $Q$ is the priority queue.

We then have

$$P_\mathcal{D} \leq P(g) \leq P_\mathcal{D} + P_Q$$

As the computation progresses, the probability mass in the queue $P_Q$ approaches zero and we get a better refinements on the value of $P(g)$. This thus forms the basis of an "anytime" algorithm for Bayesian networks.

## 5 Causation

There have been problems associated with logical formulations of causation [Pearl, 1988a]. There has been claims that Bayes networks provide the right independencies for causation [Pearl, 1988b]. This paper provides evidence that abducing to causes and making assumptions as to what to predict from those assumptions [Poole, 1989; Poole, 1990] is the right logical analogue of the independence in Bayesian networks (based on theorem 3.5).

One of the problems in causal reasoning that Bayesian networks overcome [Pearl, 1988b] is in preventing reasoning such as "if $c_1$ is a cause for $a$ and $c_2$ is a cause for $\neg a$, then from $c_1$ we can infer $c_2$". This is the problem that occurs in the Yale shooting problem [Hanks and McDermott, 1987]. Our embedding says that this does not occur in Bayesian networks as $c_1$ and $c_2$ must already be stated to be disjoint. We must have already disambiguated what occurs when they are both true. If we represent the Yale shooting scenario so that the rules for "alive" are disjoint the problem does not arise.

## 6 Comparison with Other Systems

The closest work to that reported here is by Charniak and Shimony [Charniak and Shimony, 1990; Shimony and Charniak, 1990]. Theorem 3.3 is analogous to Theorem 1 of [Shimony and Charniak, 1990]. Instead of considering abduction, they consider models that consist of an assignment of values to each random variable. The *label* of [Shimony and Charniak, 1990] plays an analogous role to our hypotheses. They however, do not use their system for computing posterior probabilities. It is also not so obvious how to extend their formalism to more powerful logics.

This work is also closely related to recent embeddings of Dempster-Shafer in ATMS [Laskey and Lehner, 1989; Provan, 1989]. One difference between our embedding of Bayes networks and Dempster Shafer is in the independence assumptions used. Dempster-Shafer assume that different rules are independent. We assume they are exclusive. Another difference is that these embeddings do not do evidential reasoning (by doing abduction), determining probability of hypotheses given evidence, but rather determine the "belief" of propositions from forward chaining.

The ATMS-based implementation is very similar to that of de Kleer and Williams [1987; 1989]. They are computing something different to us (the most likely composite hypotheses), and are thus able to do an $A*$ search. It is not clear that including the "irrelevant" hypotheses gives the advantages that would seem to arise from using an A* search.

## 7 Conclusion

This paper presented a simple but powerful mechanism for combining logical and probabilistic reasoning and showed how it can be used to represent Bayesian Networks.

Given the simple specification of what we want to compute, we are currently investigating different implementation techniques to determine which works best in practice. This includes using logic programming technology and also ATMS technology. We are also trying to the representational adequacy by building applications (particularly in diagnosis, but also in recognition), and based on this technology.

It may seem as though there is something terribly ad hoc about probabilistic Horn abduction (c.f. the extensional systems of [Pearl, 1988b]). It seems, however, that all of the sensible (where $\sum_j \mathcal{M}(a_i(v_{i,j})) = 1$ for each random variable $a_i$) representations (propositionally at least) correspond to Bayesian networks. The natural representation tends to emphasise propositional dependencies (e.g., where $b$ is an important distinction when $a$ is true, but not otherwise). These are normal Bayesian networks, but imply more structure on the contingency tables than are normally considered special.

## A  Proof Outlines of Lemmata

**Lemma 3.2** *The minimal explanations of the terminal variables having particular values correspond to the composite beliefs in the Bayesian network with the terminals having those values. The priors for the explanations and the composite beliefs are identical.*

**Proof:** First, there is a one to one correspondence between the composite beliefs and the minimal explanations of the terminals. Suppose $x_1, ..., x_n$ are the random variables such that variable $x_i$ is directly influenced by $x_{i_1}, ..., x_{i_{n_i}}$. The minimal explanations of the terminal nodes consist of hypotheses of the form

$$c\_x_i(v_i, v_{i_1}, ..., v_{i_{n_i}})$$

with exactly one hypothesis for each $x_i$, such that $x_{i_j}(v_{i_j})$ is a logical consequence of the facts and the explanation. This corresponds to the composite belief $x_1(v_1) \wedge ... \wedge x_n(v_n)$.

By construction, the proofs for the terminal nodes must include all variables.

Suppose $E$ is a minimal explanation of the terminal variables. To show there is only one hypothesis for each random variable. Suppose that $x_i$ is a variable such that there are two hypotheses

$$c\_x_i(v_i, v_{i_1}, ..., v_{i_{n_i}}), c\_x_i(v'_i, v'_{i_1}, ..., v'_{i_{n_i}})$$

in $E$. If some $x_{i_j}(v_{i_j})$ or $x_{i_j}(v'_{i_j})$ is not a consequence of $F \cup E$, then the corresponding $c\_x_i$ hypothesis can be removed without affecting the explanation, which is a contradiction to the minimality of $E$. So each $x_{i_j}(v_{i_j})$ and $x_{i_j}(v'_{i_j})$ is a consequence of $F \cup E$. By consistency of $E$ each $v_{i_j} = v'_{i_j}$. The only way these assumptions can be different is if $v_i \neq v'_i$, and so we can derive $x_i(v_i)$ and $x_i(v'_i)$ which leads to *false*, a contradiction to the consistency of $E$.

Second, the explanations and the composite beliefs have the same prior. Given an assignment of value $v_i$ to each variable $x_i$, define $\Pi_i$ by

$$\Pi_i = x_{i_1}(v_{i_1}) \wedge ... \wedge x_{i_{n_i}}(v_{i_{n_i}})$$

where $x_{i_1}, ..., x_{i_{n_i}}$ are the variables directly influencing $x_i$.

By the definition of a Bayesian net, and the definition of $c\_x_i$, we have

$$P(x_1(v_1) \wedge ... \wedge x_n(v_n))$$
$$= \prod_{i=1}^{n} P(x_i(v_i)|\Pi_i)$$
$$= \prod_{i=1}^{n} c\_x_i(v_i, v_{i_1}, ..., v_{i_{n_i}})$$
$$= P(exp)$$

Where $exp$ is the explanation. □

**Lemma 3.4** *If $H$ is a set of assignments to variables in a Bayesian Network, and $H'$ is the analogous propositions to $H$ in the corresponding Probabilistic Horn Abduction system, then*

$$P(H) = \mathcal{M}(H')$$

**Proof:** This is proven by induction on a well founded ordering over sets of hypotheses. This ordering is based on the lexicographic ordering of pairs $\langle h, n \rangle$ where $h$ is the depth of the element of the set with maximal depth, and $n$ is the number of elements of this depth. Each time through the recursion either $h$ is reduced or $h$ is kept the same and $n$ is reduced. This is well founded as both $h$ and $n$ are non-negative integers and $n$ is bounded by the number of random variables.

For the base case, where $h = 1$, we have all of the hypotheses are independent and there is only one trivial explanation. In this case we have

$$P(H) = \mathcal{M}(H') = \prod_{h \in H} P(h)$$

For the inductive case, suppose $a(v)$ is a proposition in $H$ with greatest depth. Let $R = H \setminus a(v)$. Under the ordering above $\Pi_a \cup R < H$, and so we can assume the lemma for $\Pi_a \cup R$. Note also that $a$ does not influence anything in $R$ (else something in $R$ would have greater depth than $a$).

$$
\begin{aligned}
P(H) &= P(a = v \wedge R) \\
&= P(a = v|R) \times P(R) \\
&= \left( \sum_{\Pi_a} P(a|\Pi_a \wedge R) \times P(\Pi_a|R) \right) \times P(R) \\
&= \sum_{\Pi_a} P(a|\Pi_a) \times P(\Pi_a|R) \times P(R) \\
&= \sum_{\Pi_a} P(a|\Pi_a) \times P(\Pi_a \wedge R) \\
&= \sum_{\Pi_a} P(c\_a(v, \Pi_a)) \times \mathcal{M}(\Pi'_a \cup R') \\
&= \sum_{\Pi_a} P(c\_a(v, \Pi_a)) \times \sum_{E \in expl(\Pi'_a \wedge R')} P(E) \\
&= \sum_{\Pi_a} \sum_{E \in expl(\Pi_a \wedge R)} P(c\_a(v, \Pi_a)) \times P(E) \\
&= \sum_{E' \in expl(a(v) \wedge R)} P(E') \\
&= \mathcal{M}(a(v) \wedge R') \\
&= \mathcal{M}(H')
\end{aligned}
$$

□

## Acknowledgements

This research was supported under NSERC grant OG-POO44121, and under Project B5 of the Institute for Robotics and Intelligent Systems. Thanks to Michael Horsch for working on the implementations, and to Judea Pearl for pointing out the relationship to Reichenbach's principle.